\documentclass{article}

\usepackage{hyperref}

\makeatletter
\def\@author#1{\g@addto@macro\elsauthors{\normalsize%
    \def\baselinestretch{1}%
    \upshape\authorsep#1\unskip\textsuperscript{%
      \ifx\@fnmark\@empty\else\unskip\sep\@fnmark\let\sep=,\fi
      \ifx\@corref\@empty\else\unskip\sep\@corref\let\sep=,\fi
      }%
    \def\authorsep{\unskip,\space}%
    \global\let\@fnmark\@empty
    \global\let\@corref\@empty  
    \global\let\sep\@empty}%
    \@eadauthor={#1}
}
\makeatother

\usepackage{epsfig}
\usepackage{graphicx}
\usepackage{amsmath}
\usepackage{amssymb}
\usepackage{subfig}

\title{A Generative Restricted Boltzmann Machine Based Method for High-Dimensional Motion Data Modeling}

\author{%
	Siqi Nie\footnote{Email: niesiqi@gmail.com.}\\
	Ziheng Wang \footnote{Email: wangz10@rpi.edu.}\\
	Qiang Ji\footnote{Email: jiq@rpi.edu. Affiliation: Rensselaer Polytechnic Institute, USA.}
}

\begin{document}
\maketitle

\begin{abstract}
   Many computer vision applications involve modeling complex spatio-temporal patterns in high-dimensional motion data. Recently, restricted Boltzmann machines (RBMs) have been widely used to capture and represent spatial patterns in a single image or temporal patterns in several time slices. To model global dynamics and local spatial interactions, we propose to theoretically extend the conventional RBMs by introducing another term in the energy function to explicitly model the local spatial interactions in the input data. A learning method is then proposed to perform efficient learning for the proposed model. We further introduce a new method for multi-class classification that can effectively estimate the infeasible partition functions of different RBMs such that RBM is treated as a generative model for classification purpose. The improved RBM model is evaluated on two computer vision applications: facial expression recognition and human action recognition. Experimental results on benchmark databases demonstrate the effectiveness of the proposed algorithm.
\end{abstract}

\section{Introduction}
\begin{figure}
\begin{center}
   \includegraphics[width=1\linewidth]{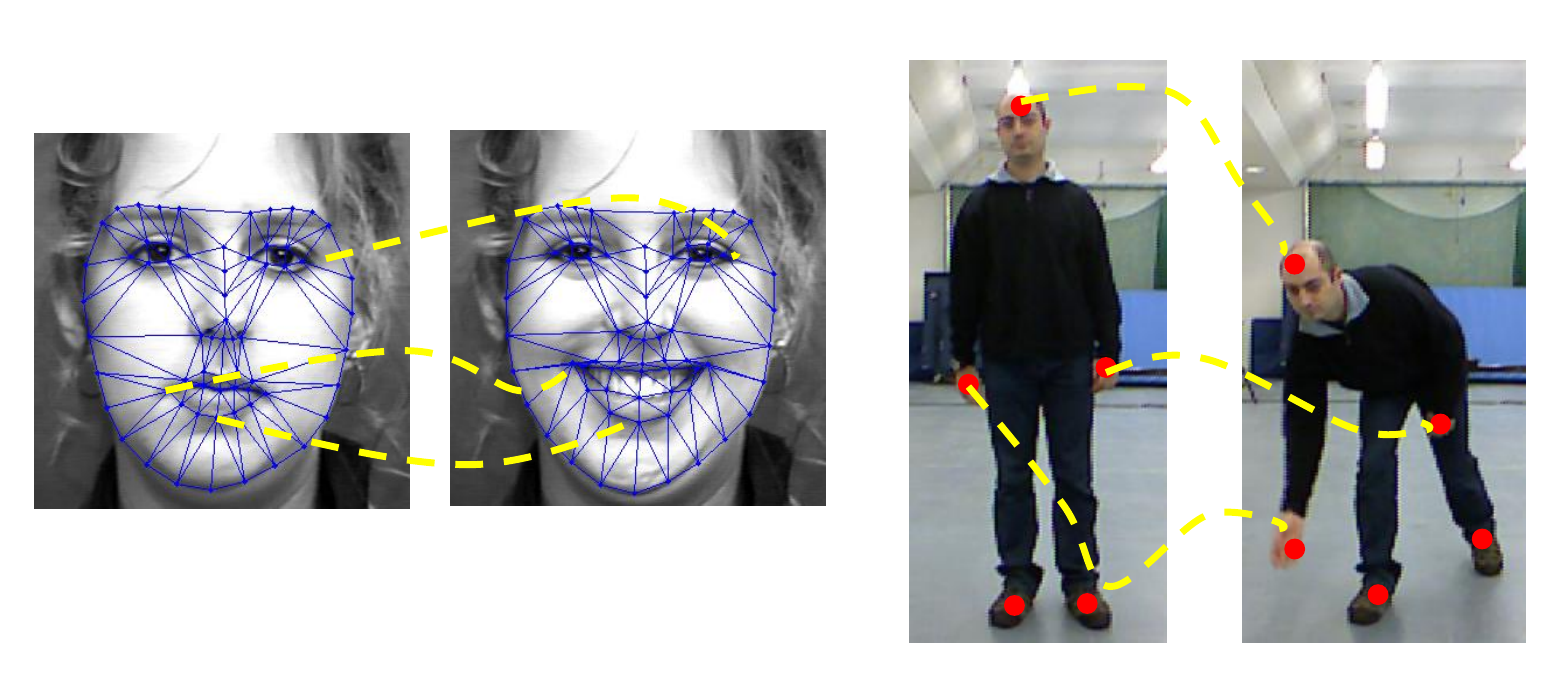}
\end{center}
   \caption{Examples of spatiotemporal patterns in different applications: \textit{facial expression recognition} and \textit{human action recognition}. Both spatial interactions and temporal movements of landmarks define an expression or an action. Images are from CK+ data set\cite{lucey2010extended} and G3D data set \cite{bloom2012g3d}, respectively.}
\label{fig:ST}
\end{figure}

Spatio-temporal patterns in high-dimensional motion data are crucial in many recognition applications. For example, human action is the combination of the body joint movements over a time interval. Facial expression is the result of the facial landmark movements (Figure \ref{fig:ST}). Understanding the movement trajectories and modeling the underlying spatio-temporal patterns play an important role in recognizing these actions, especially with the recent emergences of reliable algorithms \cite{shotton2011real, lucey2010extended} to estimate the positions of body joints and facial landmarks.

In this work, we are interested in developing a probabilistic model to capture the spatio-temporal pattenrs in high-dimensional time series for classification purpose. Many recent works develop novel models to capture the spatio-temporal dynamics \cite{laptev2005, laptev2008, lv2006recognition,  wang2012mining}. However, most models, such as hidden Markov model (HMM) \cite{lv2006recognition}, dynamic Bayesian network (DBN) \cite{dagli2003action} and conditional random field (CRF) \cite{sminchisescu2006conditional} are time-slice local models, which assume Markov property and stationary transitions and hence can only capture local dynamics. The local models suffer from two limitations. First, local dynamics may not represent a sequence well because it fails to mpdel the overall dynamics. Second, the stationary transition assumption may not hold for many real-world applications.

Compared with time-sliced dynamic models, restricted Boltzmann machines (RBMs) has shown strong capability of modeling joint distributions and therefore can capture the global patterns. In literature, RBMs have been successfully applied to separately capture the spatial \cite{eslami2012shape} or temporal \cite{wang2011anatomically} patterns in different types of data. In this work, we propose a variant of RBM that can capture spatial and global temporal patterns simultaneously to comprehensively model the high-dimensional motion data.

In a typical RBM, since there are no lateral connections among nodes in each layer, input data are independent of each other given the states of hidden layer. This assumption limits RBM's representation power, since there exist direct dependencies among input data. There are generally two types of data interactions: interactions through latent variable and direct interactions independent of latent variable. For example, as stated in \cite{osindero2007modeling}, soldiers on a parade follow the commander's order to march in some direction, but they ``form a neat rectangle by interacting with their neighbors". This example illustrates that soldiers' behavior are determined by both the commander's order (latent) and the interactions with their neighbors. In this case, RBM is not effective in modeling the local interaction that is independent of the latent variable. The use of RBM for data representation and classification is further hindered by the difficulty in comparing different RBMs due to the intractable computation of the partition functions.

Allowing interaction among visible units can overcome this shortcoming. We introduce restricted Boltzmann machine with local interactions (LRBM) to capture both the global temporal patterns and local spatial interactions in the input data. Specifically, we add a new pairwise potential term in the learning objective function of RBM to capture the local spatial interactions among components of an input vector. To perform efficient learning for the model, we replace the reconstruction procedure in the conventional Contrastive Divergence \cite{carreira2005contrastive} algorithm with a mean field approach. 

For classification task, RBM is typically used for learning features, as the input to a second stage classifier (e.g., SVM \cite{lee2009convolutional}). Typically one model is trained for all classes. To obtain good features, deep structure is built and back propagation is employed to carefully tune the parameters. In this work, we use RBM as a generative model to capture the spatio-temporal patterns in the data. RBM is used for data representation instead of feature learning. Given an observation, the only output we get from an RBM model is the likelihood of the model. For the recognition purpose, one model is trained for one class of input data. To compare among different models, we employ a method to estimate the relative partition function of a pair of RBMs for binary classification, and a label ranking method is used to extend the binary classification to multi-class classification.


To evaluate the performance of LRBM, we apply it to two areas related to complex spatial and temporal patterns: facial expression recognition and human action recognition. Experimental results on benchmark databases demonstrate the effectiveness of the proposed model.

The rest of the paper is structured as follows. Section \ref{related work} presents an overview of the related work. Section \ref{RBM} introduces the LRBM to model the global temporal dynamics and spatial patterns of motion data, as well as the classification method. We will then give the experimental results in \ref{experiments}. The paper is concluded in section \ref{conclusion}.

\section{Related Work}
\label{related work}

Capturing and representing spatio-temporal structure in data is important for many recognition and classification tasks. Research for capturing such patterns can be categorized into feature-based and model-based methods. The most widely used spatio-temporal features include spatio-temporal interest point (STIP) based features \cite{laptev2005, laptev2008} and optical flow based features \cite{cutler1998view}. These features capture local appearance or motion patterns near the interest points or optical flows. Although having been successfully applied to many applications, these features generally focus more on local patterns.

Model-based methods include probabilistic graphical models such as Hidden Markov Models \cite{lv2006recognition}, Dynamic Bayesian Networks \cite{yang2010driver}, Conditional Random Fields \cite{packer2012combined}, and their variants. While capable of simultaneously capturing both spatial and temporal interactions, they can only capture the local spatial and temporal interactions due to the underlying Markov assumption.

Restricted Boltzmann machines (RBMs) have been separately used for modeling spatial correlation or temporal correlation in the data in the last decade. RBM was firstly introduced to learn deep features from handwritings to recognize digits \cite{hinton2006reducing}. In \cite{eslami2012shape}, Eslami et al. propose a Deep Belief
Network to model the shapes of horses and motorbikes. The samples from the model look realistic and have a good generalization.
A more complicated model, proposed by Nair and Hinton \cite{nair20093}, considers the spatial correlation among visible layer using a factored 3-way RBM, in which triple potentials are used to model the correlations among pixels in natural images. The intuition is that in natural images, the intensity of each pixel is approximately the average of its neighbors. Wu et al. \cite{wu2013RBM} apply the 3-way RBM to facial landmark tracking, and model the relationship between posed faces and frontal faces, under varying facial expressions.

For dynamic data modeling, Taylor et al. \cite{taylor2007modeling} use a Conditional RBM (CRBM) to model the temporal transitions in human body movements, and reconstruct body movements. Nevertheless, like HMM, CRBM still models local dynamics by assuming $n$'th order Markov property. 

The idea of using RBM to model global pattern is not new. In \cite{kae2013augmenting, kae2014shape}, RBM and CRF are combined for face labeling problem in a simgle image or video sequences. Compared with these works, our work is different for several reasons. First, our goal is to use RBM for data representation for  multi-class classification, while the goal of \cite{kae2013augmenting, kae2014shape} is MAP inference, which is to recover the label for each superpixel. Second, we extend the conventional RBM to capture the local shape and global temporal patterns in a unified model, while in \cite{kae2013augmenting, kae2014shape} RBM is built on top of the hidden layer of the CRF, as the prior for the labels. Third, our method learns all the parameters simultaneously, while their method performs learning separately.

RBM and its variants have also been used for modeling motion data. For example, Sutskever and Hinton \cite{sutskever2007learning} introduce a temporal RBM to model high-dimensional sequences. Wang et al. \cite{wang2011anatomically} use RBM to capture the global dynamics of finger trace. However, it is limited to model the global dynamics for 1-D data only.  

To improve the representation power of RBM, semi-restricted Boltzmann machine (SRBM) \cite{osindero2007modeling} is introduced to model the lateral interactions between visible variables. The main property of SRBM is that given the hidden variables, the visible layer forms a Markov random field. However, for high-dimensional motion data, there will be too many parameters if every pair of visible units has an interaction. In this work, we model the dynamic nature of data with fewer parameters than a SRBM, which makes the learning process more efficient.

Besides feature extraction and shape modeling, RBMs have also been used for classification. Larochelle and Bengio \cite{larochelle2008classification} introduce a discriminative RBM as a classifier by including the labels in visible layer, and make predictions by comparing the likelihood of each label vector. In \cite{laro2011} discriminate RBM is introduced to model vector inputs by duplicating the discriminative RBM and adding constraints on the hidden layer. In all RBM related models discussed above, one RBM is trained for all classes. For this work, in contrast, we build one RBM for each class, and perform a multi-class classification task.

In this research, we propose to extend the standard RBM to simultaneously capture the spatial and global temporal patterns in high-dimensional sequential data 
and employ such RBM model to different classification tasks in computer vision.

\section{Data Spatio-Temporal Pattern Modeling}
\label{RBM}
In this section, we will firstly give a brief introduction of restricted Boltzmann machine, and then introduce the proposed RBM with Local Interaction (LRBM) to model multi-dimensional motion data and its learning method. Finally we present the method to use LRBMs as a set of pairwise classifier to perform multi-class recognition.

\subsection{Restricted Boltzmann Machine}

\begin{figure}
\begin{center}
   \includegraphics[width=0.5\linewidth]{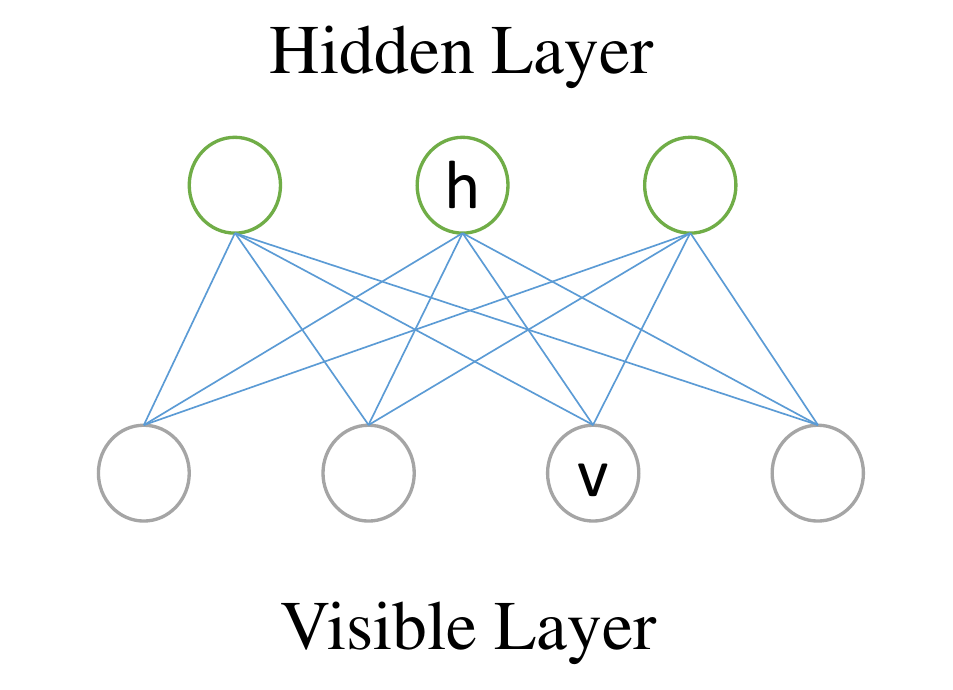}
\end{center}
   \caption{Illustration of a standard RBM}
\label{fig:RBM}
\end{figure}

A standard RBM is a generative model with two densely connected layers, one visible layer to represent data and one latent layer to extract stochastic binary features from data (Figure \ref{fig:RBM}). Hidden units are connected to visible nodes using symmetrically weighted connections to model their joint distribution. In our work, we use Gaussian-Binary RBM, where the hidden units are binary and the visible variables are assumed to follow normal distribution. 

The energy function $E(\mathbf{v},\mathbf{h})$ for each pair $(\mathbf{v}, \mathbf{h})$  is parameterized in Equation \ref{RBM energy function}. 
\begin{equation}
E\left(\mathbf{v},\mathbf{h}\right)=\sum_{i}\frac{\left(v_{i}-a_{i}\right)^2}{2\sigma_{i}^2}-\sum_{i,j}\frac{v_{i}}{\sigma_{j}}w_{ij}h_{j}-\sum_j b_{j}h_{j}\, ,
\label{RBM energy function}
\end{equation}
where $a_{i}$ is the bias for visible unit $v_{i}$, $\sigma_{i}$ is the standard deviation of the Gaussian distribution, which is typically 1 if we normalize the data, $b_{j}$ is the bias for the hidden unit $h_{j}$, $w_{ij}$ is the weight of the link connecting $v_i$ and $h_j$.

For every possible pair of visible and hidden vector, the network assigns a probability:
\begin{equation}
p\left(\mathbf{v},\mathbf{h}\right)=\frac{1}{Z}\exp\left(-E\left(\mathbf{v},\mathbf{h}\right)\right)\, ,
\label{RBM joint distribution}
\end{equation}
where $Z$ is the partition function. With continuous inputs, the $Z$ is calculated by integrating over all visible nodes and summing over all hidden unit configurations.

\subsection{Proposed Model}

Using standard RBM to model high-dimensional motion data has its limitations. First, the interaction among the data is represented through latent variables, which can be easily represented in direct connections. Second, if a single RBM models the whole sequence, the spatial information in a time slice is treated the same as the temporal information of one dimension. If one RBM models only one time slice (as in \cite{taylor2007modeling}), the temporal information remains local.

\begin{figure}
\begin{center}
   \includegraphics[width=0.75\linewidth]{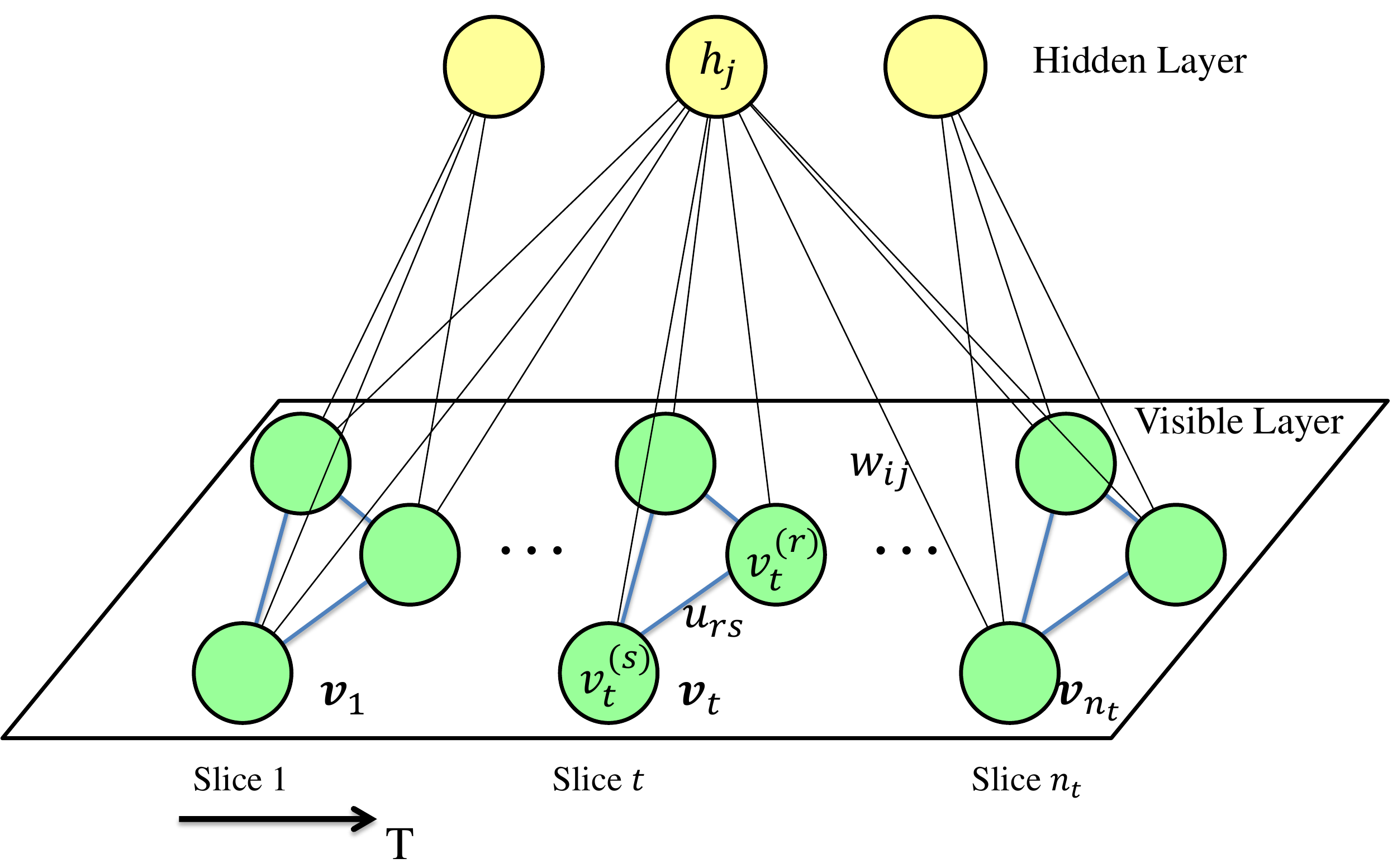}
\end{center}
   \caption{Illustration of the proposed LRBM model. Each hidden unit connects to each visible unit to capture the global temporal patterns. (Some links are omitted for brevity.) Connections within each time slice are to capture spatial pattern. The local covariance is parameterized using symmetric matrix $\mathbf{U}$, which is consistent over time.}
\label{fig:LRBM}
\end{figure}

In this work, we propose the restricted Boltzmann machine with local interaction (LRBM, Figure \ref{fig:LRBM}) to overcome such limitations. For $d \times n_t$ sequential data $\mathbf{V}=[\mathbf{v}_1, \mathbf{v}_2, \dots , \mathbf{v}_{n_t}]$, by allowing local interaction, we have the following energy function,
\begin{eqnarray}
E(\mathbf{V},\mathbf{h})=\frac{1}{2} \sum_{i=1}^{n_t} (\mathbf{v}_i-\mathbf{a}_i)^T (\mathbf{v}_i-\mathbf{a}_i)-\sum_{j=1}^{n_h}b_j h_j\nonumber\\
  -\sum_{i=1}^{n_t} \sum_{j=1}^{n_h} \mathbf{v}_i^T \mathbf{w}_{ij\cdot} h_j - \frac{1}{2}\sum_{i=1}^{n_t}\mathbf{v}_i^T\mathbf{U}\mathbf{v}_i\, ,
\label{eqn: energy}
\end{eqnarray}
where $\mathbf{v}_i$ is a $d$-dimension vector representing input vector at time slice $i$, $\mathbf{w}$ has the dimension of $n_t \times n_h \times d$, $\mathbf{w}_{ij\cdot}$ is the weight vector connecting $\mathbf{v}_i$ to a hidden node $h_j$. $\mathbf{a}_i$ and $b_j$ have the same meaning as in Equation \ref{RBM energy function}. $\mathbf{U}$ is a $d \times d$ symmetric matrix with zeros on diagonal, modeling the correlation of each vector $\mathbf{v}_i$. 

The proposed energy function models two kinds of data interactions: interaction through latent variables and interaction directly among data. In high-dimensional motion data, components in a single time slice (spatial information) is better to be considered differently from components along the timeline (temporal information). Interactions between input data and hidden variables are through weight $\mathbf{w}$, which models global pattenrs, because every visible unit is connected to every latent unit. Matrix $\mathbf{U}$ models direct interactions among input data, as in Figure \ref{fig:LRBM}, representing local spatial patterns. This kind of interaction directly affects visible layer without going through latent layer, so it is more effective to model some local spatial patterns. The elements in $\mathbf{U}$ are pairwise potentials of features in one time slice. Different shapes (spatial pattern) of input will have different contributions to the energy function. Thus, $\mathbf{U}$ captures spatial patterns among elements of input vector. With temporal information captured via the hidden nodes, our work can capture global spatio-temporal dynamics.

We assume the spatial relationship are constant throughout the whole sequence, so the parameters in $\mathbf{U}$ are shared in different frames, which means $\mathbf{U}$ is invariant of $i$. Under this invariance assumption of $\mathbf{U}$, the number of parameters is significantly reduced.

From the energy function (Equation \ref{eqn: energy}) and likelihood function (Equation \ref{RBM joint distribution}), we can derive the probability of an hidden unit to be activated, given an visible layer:
\begin{equation}
p(h_j=1|\mathbf{V})=\text{sigmoid}(b_j + \sum_i \mathbf{v}_i^T \mathbf{w}_{ij\cdot})\, .
\label{eqn: prob hidden}
\end{equation}

As one hidden unit is connected to all visible units, whether it is activated or not depends stochastically on the visible layer. A hidden unit $h_{j}$ is activated through Equation \ref{eqn: prob hidden} when it detects some specific pattern in the visible layer. The pattern is captured by the weights connecting each element in $\mathbf{V}$ to $h_{j}$. Therefore the hidden layer $\mathbf{h}$ is able to capture important global patterns of $\mathbf{V}$. Hence, the proposed model can simultaneously capture the global temporal patterns (through $\mathbf{w}$) and local shape patterns (through $\mathbf{U}$).

LRBM is similar to the factored 3-way RBM \cite{nair20093} and SRBM \cite{osindero2007modeling} with respect to modeling the interactions in the visible layer. However, there exist some significant differences. First, in factored 3-way RBM, every pair of visible units has a potential to represent the interaction, while in LRBM, units only interact with neighbors in the time slice. Second, if the number of factors in 3-way RBM or the data dimension in SRBM is high, the overall parameters are much more than LRBM. In LRBM, the additional parameters are from matrix $\mathbf{U}$, with the $d(d-1)/2$ parameters, where $d$ is the dimension of the vector $\mathbf{v}_i$. This would greatly reduce the computational load. Finally, both 3-way RBM and semi-RBM are proposed to model the spatial patterns of a single natural image, while LRBM models spatio-temporal patterns of sequential data.

The joint probability of $\mathbf{V}$ and $\mathbf{h}$ are the same as in Equation \ref{RBM joint distribution}. The likelihood of input data is computed by summing over all configurations of hidden units:
\begin{equation}
p(\mathbf{V})=\frac{1}{Z}\sum_{\mathbf{h}}e^{-E(\mathbf{V},\mathbf{h})}\, .
\label{eqn: likeli}
\end{equation}

\subsection{Model Learning}
In our work, one LRBM is trained for one class. The parameters include bias for visible and hidden units $\mathbf{a}_i$ and $b_j$, weight between two layers $\mathbf{w}$, and local interaction matrix $\mathbf{U}$. To learn the parameters, we seek to maximize the joint probability of all training data $D$ of a class, $P(D)=\prod_{\mathbf{V} \in D}p(\mathbf{V})$. Assume the data has been normalized, so $\mathbf{a}_i$ can be removed from the energy function. The derivative of the log-likelihood of a training instance with respect to a parameter is given below:
\begin{eqnarray}
\frac{\partial \log p(\mathbf{V})}{\partial \mathbf{w}_{ij}}&=&\langle \mathbf{v}_i h_j\rangle_{data}-\langle \mathbf{v}_i h_j\rangle_{model}\, ,\\
\frac{\partial \log p(\mathbf{V})}{\partial b_{j}}&=&\langle h_j\rangle_{data}-\langle h_j\rangle_{model}\, ,\\
\frac{\partial \log p(\mathbf{V})}{\partial u_{rs}}&=&\langle \sum_i^{n_t}\mathbf{v}_i^{(r)} \mathbf{v}_i^{(s)}\rangle_{data}\nonumber\\&&-\langle \sum_i^{n_t}\mathbf{v}_i^{(r)} \mathbf{v}_i^{(s)}\rangle_{model}\, ,
\end{eqnarray}
where the angle brackets are used to denote expectations under the distribution specified by the subscript that follows. $u_{rs}$ is the element of $\mathbf{U}$ at position $(r, s)$. $\mathbf{v}_i^{(r)}, \mathbf{v}_i^{(s)}$ are the $r^{\text{th}}$ and $s^{\text{th}}$ components of vector $\mathbf{v}_i$. This leads to a simple learning rule: stochastic steepest ascent in the log-likelihood of training data. Take $\mathbf{w}_{ij}$ as an example:
\begin{equation}
\Delta \mathbf{w}_{ij} = \epsilon (\langle \mathbf{v}_i h_j\rangle_{data}-\langle \mathbf{v}_i h_j\rangle_{model})\, ,
\end{equation}
where $\epsilon$ is the learning rate. $\langle \mathbf{v}_i h_j\rangle_{data}$ is easy to get from the data. $\langle \mathbf{v}_i h_j\rangle_{model}$ is much more difficult to compute due to the large dimension of hidden and visible layers. Hinton \cite{hinton2002training} proposes the CD algorithm to approximate $\langle \mathbf{v}_i h_j\rangle_{model}$ by using a reconstructed sample, which gives the following weight change: 
\begin{equation}
\Delta \mathbf{w}_{ij} = \epsilon (\langle \mathbf{v}_i h_j\rangle_{data}-\langle \mathbf{v}_i h_j\rangle_{recon})\, .
\end{equation}

Given the visible layer, the hidden units are independent with each other, so sampling the states of hidden units can be performed in parallel using Equation \ref{eqn: prob hidden}.

Given the states of the hidden units, the visible units form a Markov Random Field in which the pairwise interaction weight between $\mathbf{v}_i^{(r)}$ and $\mathbf{v}_i^{(s)}$ is $u_{rs}$. They are no longer independent, so sampling cannot be done in parallel. However, we can obtain the conditional probability of each node by fixing its neighbors. The mean field algorithm is used to sample data from the model, and to reconstruct visible layer when learning the parameters of LRBM. Specifically, for each vector $\mathbf{v}_i$, each component $\mathbf{v}_i^{(s)}$ is sampled by fixing all its neighbors. Once a component is sampled, the vector is updated with the newly sampled component. This procedure is repeated until all components have been updated, and thus we get a reconstructed vector. The conditional probability of one visible node is given in Equation \ref{eqn: prob visible}:
\begin{equation}
p(\mathbf{v}_i^{(s)}|\text{N}(\mathbf{v}_i^{(s)}), h)=
\mathcal{N}(\mathbf{a}_i^{(s)}+\sum_j h_j \mathbf{w}_{ij\cdot}^{(s)}+\sum_{v_k\in \text{N}(\mathbf{v}_i^{(s)})}v_k u_{ks}, 1)\, ,
\label{eqn: prob visible}
\end{equation}
where $\text{N}(\mathbf{v}_i^{(s)})$ are the neighbors of $\mathbf{v}_i^{(s)}$, $\mathcal{N}(\mu, \sigma^2)$ denotes the Gaussian probability density function with mean $\mu$ and variance $\sigma^2$. The superscript $^{(s)}$ means the $s^{\text{th}}$ component of a vector.

Compared with distribution of visible units in standard RBM:
\begin{equation}
p(v_i|h)=\mathcal{N}(a_i+\sum_j h_j w_{ij}, 1)\, ,
\end{equation}
the mean of the distribution of a visible node is similar in the first two terms, except that in LRBM, it is modified by the pairwise potentials relating one node to all its neighbors.


With the reconstructed data, we are able to calculate the approximate gradient of each parameter, and perform the CD learning procedure. As mentioned in \cite{hinton2010practical}, the RBMs learn better if more steps of reconstruction are used before collecting the statistics. In practice, we sample 10 times using the mean field method before collecting the reconstructed data in each epoch. 


Due to the non-convex property of the objective function, different parameter initializations in RBM learning can end up with different models. For each class, several candidate models are learned from different initializations, and we select the one that can best discriminate current class from the others. 

For example, after learning model $\mathcal{M}_{1}$ for class $\mathcal{C}_{1}$, given instances from all classes $\{I_1, I_2, \cdots, I_N\}$, we expect instance $I_1$ to have greater likelihood on model $\mathcal{M}_{1}$ than all other instances ($\{I_2, \cdots, I_N\}$). In practice, we can sort the likelihoods of instances from all classes, and choose the candidate model that minimizes the rank of the instances from class $\mathcal{C}_{1}$.

Notice that the likelihood is calculated according to Equation \ref{eqn: likeli}, given a single LRBM model, so for the comparison of the likelihoods, we don't need to estimate the intractable partition function, since it is merely a constant.

\subsection{Multi-class Classification}
\label{RBM for classification}
For this work, RBM is used as a generative model for classification. A common strategy for training generative model for classification is to train multiple models for different classes and then evaluate the likelihood of each model during testing. In particular, given a set of properly learned LRBM's $\{\mathcal{M}_{i}\}_{i=1}^{N}$ for $N$ classes, the basic idea for classification is to find the model that generates the largest likelihood given an instance of data $\mathbf{V}$.
\begin{equation}
i^{*} = \arg\max_{i}p(\mathbf{V}|\mathcal{M}_i)\, ,
\end{equation}
where $i^{*}$ is the prediction of our classifier. The likelihood of a data instance is
\begin{eqnarray}
\log p(\mathbf{V}) = -\frac{1}{2} \sum_{i=1}^{n_t} (\mathbf{v}_i-\mathbf{a}_i)^T (\mathbf{v}_i-\mathbf{a}_i) + \sum_{i=1}^{n_t} \mathbf{v}_i^T \mathbf{U} \mathbf{v}_i \nonumber\\
+ \sum_{j=1}^{n_h}\log\left(1+\exp\left(b_j+\sum_{i=1}^{n_t}\mathbf{v}_i^T \mathbf{w}_{ij}\right)\right) - \log Z\, .
\label{eqn: detailed loglik}
\end{eqnarray}

For brevity, we denote $p(\mathbf{V})$ as
\begin{equation}
p(\mathbf{V})=g(\mathbf{V})-\log Z\, .
\end{equation} 

$g(\mathbf{V})$ can be computed directly. However the partition functions $Z$ are intractable for RBMs with large numbers of hidden and visible units. One possible solution is the Annealed Importance Sampling \cite{salakhutdinov2008quantitative} to estimate the partition functions, and directly compare the likelihood. However, such estimation needs too many samples to obtain a good estimation of the partition function. Instead, for binary classification, Schmah et al. \cite{schmah2008generative} propose a method to discriminatively estimate the difference of log-partition functions of two RBMs:

\begin{equation}
c_{ij}=\log Z_i-\log Z_j\, .
\end{equation}

We extend this method to multi-class classification. Ideally, $c_{ij}=c_{ik}+c_{kj}$. But since the partition function is not directly computed, there is a slight error in the estimation. To address this issue, we perform a label ranking process \cite{hullermeier2008label} to make the final decision using a set of pairwise classifiers.

In the simplest case, with all the pairwise classifiers $f_{ij}$, each prediction is interpreted as a vote for a class, and the class with the highest votes is proposed as a prediction. Alternatively, in confidence estimation, instead of a binary result \{0, 1\}, a ``soft" classifier is employed to map the difference in likelihood into the unit interval [0, 1]:
\begin{equation}
\mathcal{F}_{ij}(\mathbf{V})=\frac{1}{1+\exp(-\alpha (g(\mathbf{V}|\mathcal{M}_i)-g(\mathbf{V}|\mathcal{M}_j)-c_{ij}))}\, .
\end{equation} 

Parameter $\alpha$ is searched in a range of $[0.01, 100]$ to maximize the training accuracy.

The output of such ``soft" binary classifier can be interpreted as a confidence value in the classification: the closer the output $\mathcal{F}_{ij}$ to 1, the stronger the decision of choosing class $\mathcal{C}_i$ is supported. Notice that $\mathcal{F}_{ij}$ is not symmetrical. It only gives the preference toward $i$ between $i$ and $j$. The preference toward $j$ is $1-\mathcal{F}_{ij}$. A valued preference relation matrix $\mathcal{R}_{\mathbf{V}}$ is defined for any query instance $\mathbf{V}$:
\begin{equation}
\mathcal{R}_{\mathbf{V}}(i, j) = \left\{ 
  \begin{array}{l l}
    \mathcal{F}_{ij}(\mathbf{V}) & \quad if \quad i<j\\
    1-\mathcal{F}_{ij}(\mathbf{V}) & \quad if \quad i>j\\
  \end{array} \right. \, .
\label{preference relation}
\end{equation}

In our approach, we evaluate the confidence score by summing up all the confidence value
\begin{equation}
\mathcal{S}_{\mathbf{V}}(i)=\sum _{j\neq i}\mathcal{R}_{\mathbf{v}}(i, j)\, ,
\label{score}
\end{equation}
where the index $i$ goes through $1$ to $N$, meaning the confidence score for one instance on different classes. The label of the model that associates with the highest score is proposed as the final decision.
\begin{equation}
i^*=\arg\max_i \mathcal{S}_\mathbf{V}(i)\, .
\end{equation}

In general, if we have a $N$-class classification problem, ${n \choose 2}=n(n-1)/2$ pairwise classifiers are used to compute the confidence score.

\section{Experiments}
\label{experiments}
We evaluate our algorithm on two different areas that involve high-dimensional motion data: facial expression recognition and human action recognition. Two benchmark data sets are used in our experiment: the extended Cohn-Kanade data set (CK+) \cite{lucey2010extended} and G3D data set \cite{bloom2012g3d}. We will also compare the proposed LRBM model with other related methods. 

\subsection{Facial Expression Recognition}

The Extended Cohn-Kanade data set (CK+) is a complete data set for action units and emotion-specified expressions. In this work, based on a sequence of landmarks on the human face, our goal is to classify it into one of seven expressions: \textit{angry, disgust, fear, happy, sadness, surprise} and \textit{contempt}. CK+ data set includes 593 sequences from 123 subjects. From all the sequences, 327 are associated with expression labels. In our experiment, these 327 sequences are used. Landmarks are the positions of 68 facial points from a detection and tracking procedure, which are provided by the database. 

\begin{figure}
\begin{center}
   \includegraphics[width=0.5\linewidth]{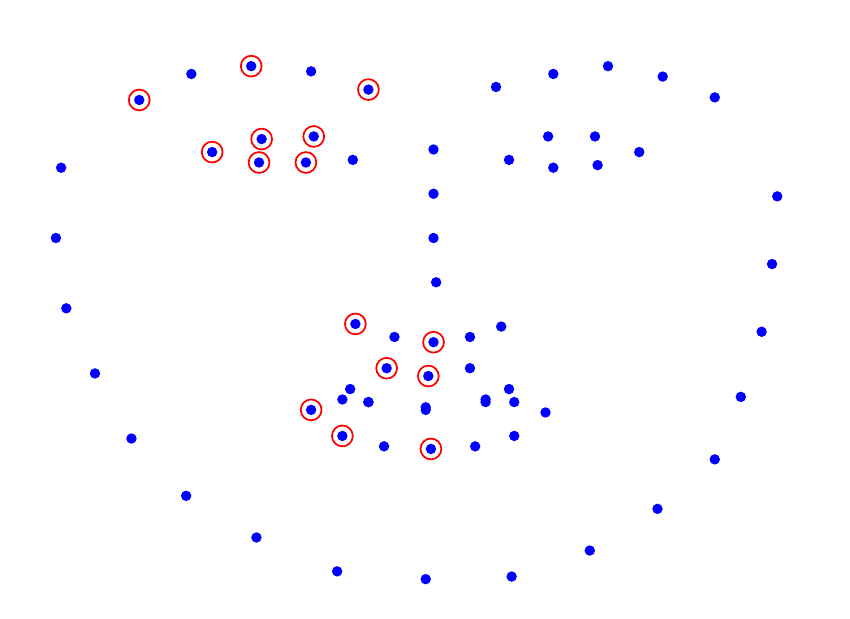}
\end{center}
   \caption{The 15 points (circled dots) we selected as the input data for each time slice. Because of facial symmetry, we only select points on the left side of the face, and points along the vertical center line.}
\label{fig:featureck}
\end{figure}

As the pose of each subject varies slightly over time, we apply a pose rectification procedure to make each face a frontal one. Then, using the detected eyes and the interocular distance, we perform a geometric normalization by making the size of facial area the same throughout all subjects, and moving the centers of eyes to the same position. A smoothing filter is also used to reduce the noise in the trajectory. To reduce the feature dimension while keeping the useful information, as shown in Figure \ref{fig:featureck}, we omit the outline landmarks and some other points, and select the more informative points, resulting in a 15-point feature for each frame, which is a 30d vector. The features selected are reasonable due to the property of facial symmetry.

\begin{table}
\begin{center}
\caption{Confusion Matrix of the classification performance of LRBM on CK+ data set (\%). For brevity, \textit{An, Di, Fe, Ha, Sa, Su, Co} correspond to \textit{angry, disgust, fear, happy, sadness, surprise, contempt}, respectively.}
\label{Table: result exp}
\begin{tabular}{|c|c|c|c|c|c|c|c|}
\hline
 & An & Di & Fe & Ha & Sa & Su & Co \\
\hline
An & 97.8 & 0.0  & 0.0  & 0.0  & 0.0  & 2.2  & 0.0 \\
Di & 8.5  & 89.8 & 0.0  & 1.7  & 0.0  & 0.0  & 0.0 \\
Fe & 0.0  & 0.0  & 84.0 & 8.0  & 0.0  & 8.0  & 0.0 \\
Ha & 0.0  & 0.0  & 0.0  & 100.0 & 0.0  & 0.0  & 0.0 \\
Sa & 10.7  & 3.6  & 3.6  & 0.0  & 78.6 & 3.6  & 0.0 \\
Su & 0.0 & 0.0  & 0.0  & 0.0  & 1.2  & 97.6 & 1.2 \\
Co & 5.6  & 0.0  & 0.0  & 11.1  & 11.1  & 0.0  & 72.2 \\
\hline
\end{tabular}
\end{center}
\end{table}

\subsubsection{Classification Performance}

Intuitively, the spatial patterns are crucial in recognizing an expression. From a single face, human can identify the expression without any difficulty. But in order to explore the underlying facial dynamics as additional features for our system, we choose 10 frames from each sequence, representing the neutral look, intermediate expressions and the peak expression. Since our method captures global temporal patterns, a good alignment of different sequences is important to ensure recognition performance. Linear interpolation is used to get sequences with fixed length. The size of hidden layer is set as 400. To compare with baseline method, we use a leave-one-subject-out cross-validation configuration. Each time we generate the testing data from sequences of one subject. 25 other subjects form the validation set, and all the other subject form the training set.

\begin{table}
\begin{center}
\caption{Classification accuracy of LRBM and other state-of-the-art approaches (\%)}
\label{Table: Comparison}
\begin{tabular}{|c|c|c|c|}
\hline
 & Lucey et al. \cite{lucey2010extended} & Wang et al. \cite{wang2012} & LRBM \\
\hline\hline
An & 75.0 & 91.1 & \textbf{97.8} \\
Di & \textbf{94.7} & 94.0 &  89.8 \\
Fe & 65.2 & 83.3 &  \textbf{84.0} \\
Ha & \textbf{100.0} & 89.8 & \textbf{100.0} \\
Sa & 68.0 & 76.0  & \textbf{78.6} \\
Su & 96.0 & 91.3  & \textbf{97.6} \\
Co & \textbf{84.4} & 78.6  & 72.2 \\\hline
Avg. & 83.3 & 86.3 & \textbf{88.6} \\\hline
\end{tabular}
\end{center}
\end{table}

The performance of the algorithm is given in Table \ref{Table: result exp}. As we only use the shape features (i.e. facial landmark positions), the comparison is only between methods using the same features. Our method's hit rates for each expression are: Angry - 97.8\%, Disgust - 89.8\%, Fear - 84.0\%, Happy - 100.0\%, Sadness 78.6\%, Surprise - 97.6\%, Contempt - 72.2\%. The average accuracy of all the expressions is 88.6\%, which is more than 2\% better than state-of-the-art method, as reported in \cite{wang2012}. In Table \ref{Table: Comparison}, we list the comparison of our model with some recent approaches. Our approach reaches the best accuracy on five expressions (\textit{angry, fear, happy, sadness and surprise}) among all the methods listed. Although our model does not achieve the best accuracy for every expression, it does not fall behind very much on \textit{Disgust}, but it fails sometimes on \textit{Contempt}, because this is a subtle expression that our method needs more features to represent it accurately. In CK+ data set, contempt expression has only 18 samples. However, the average performance of the proposed method has been improved a lot. Another thing to mention is that the method in \cite{lucey2010extended} uses both shape features and appearance features, while we only use the shape features.


If we do not consider the spatial interaction within each frame, the data is simply aligned as a long vector to feed in the conventional RBM. Then the performance is 86.3\%, which is less than the LRBM. This also proves the improvement of LRBM over RBM.

\begin{figure}
  \centering
  \subfloat[]
  {
  \includegraphics[width=0.32\textwidth]{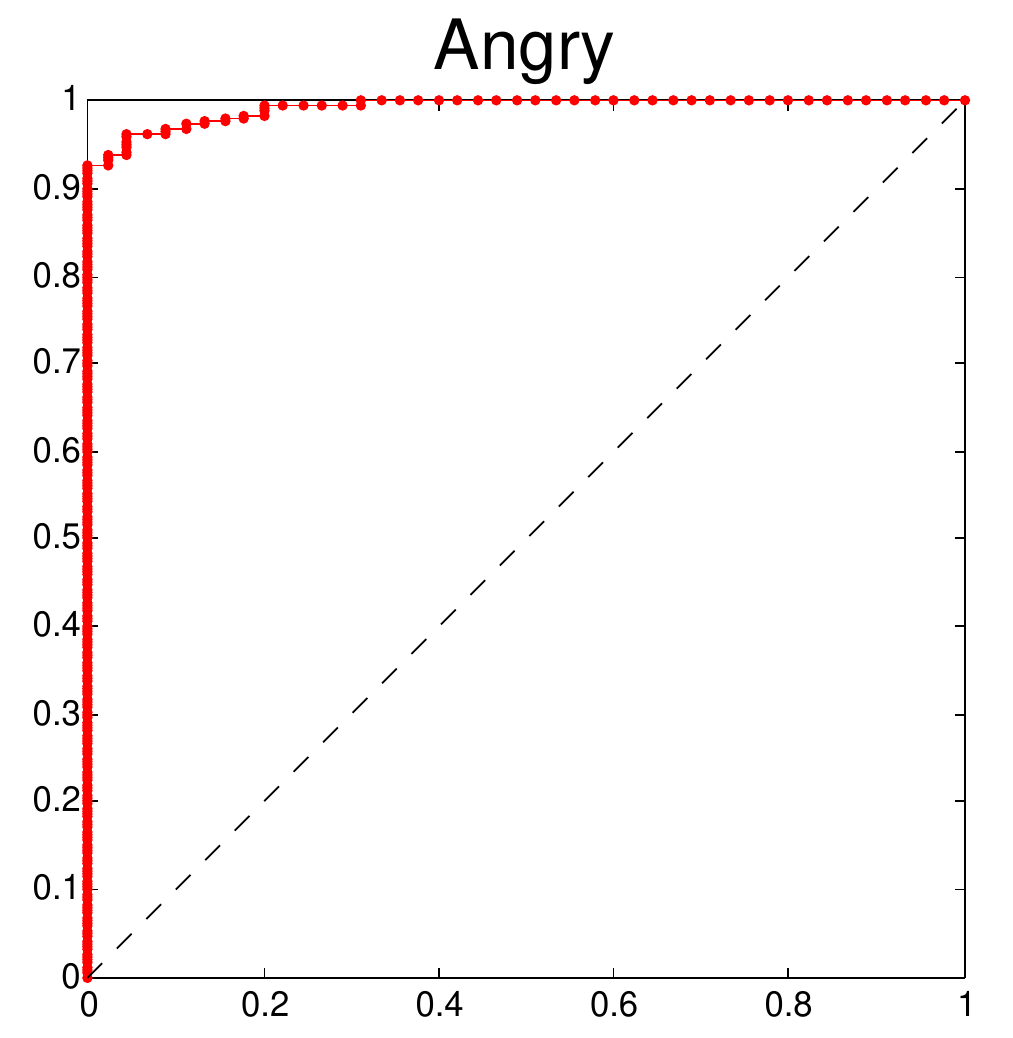} }        
  \subfloat[]
  {
  \includegraphics[width=0.32\textwidth]{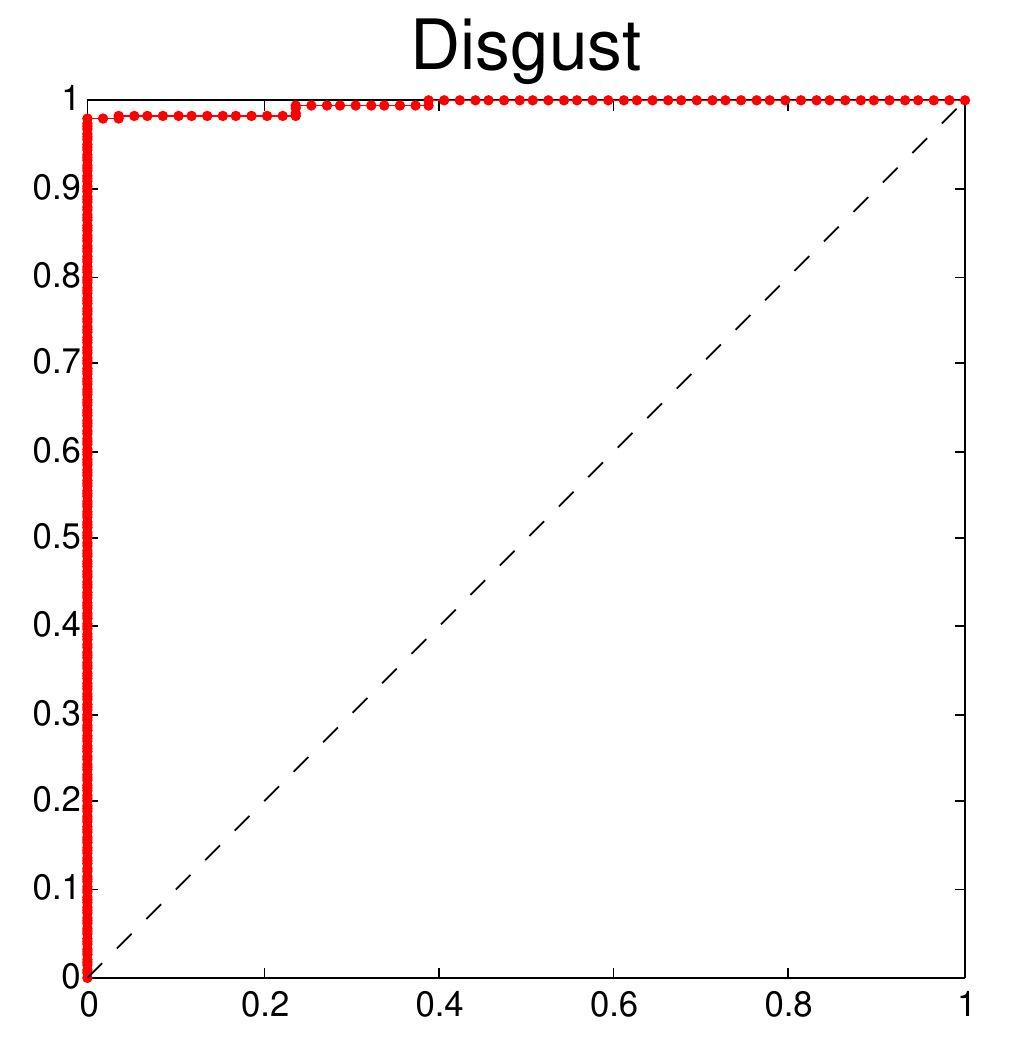} }
  \subfloat[]
  {
  \includegraphics[width=0.32\textwidth]{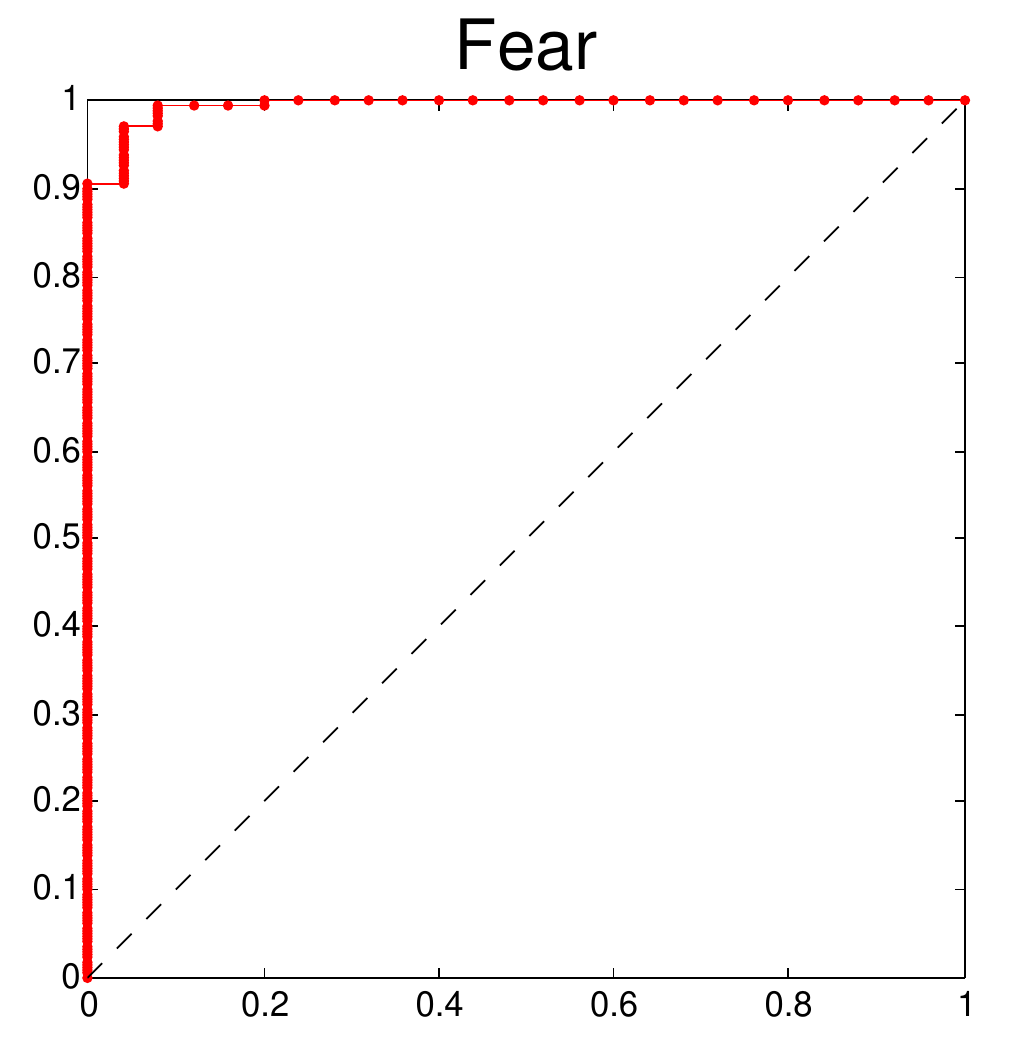} }\\
  \subfloat[]
  {
  \includegraphics[width=0.32\textwidth]{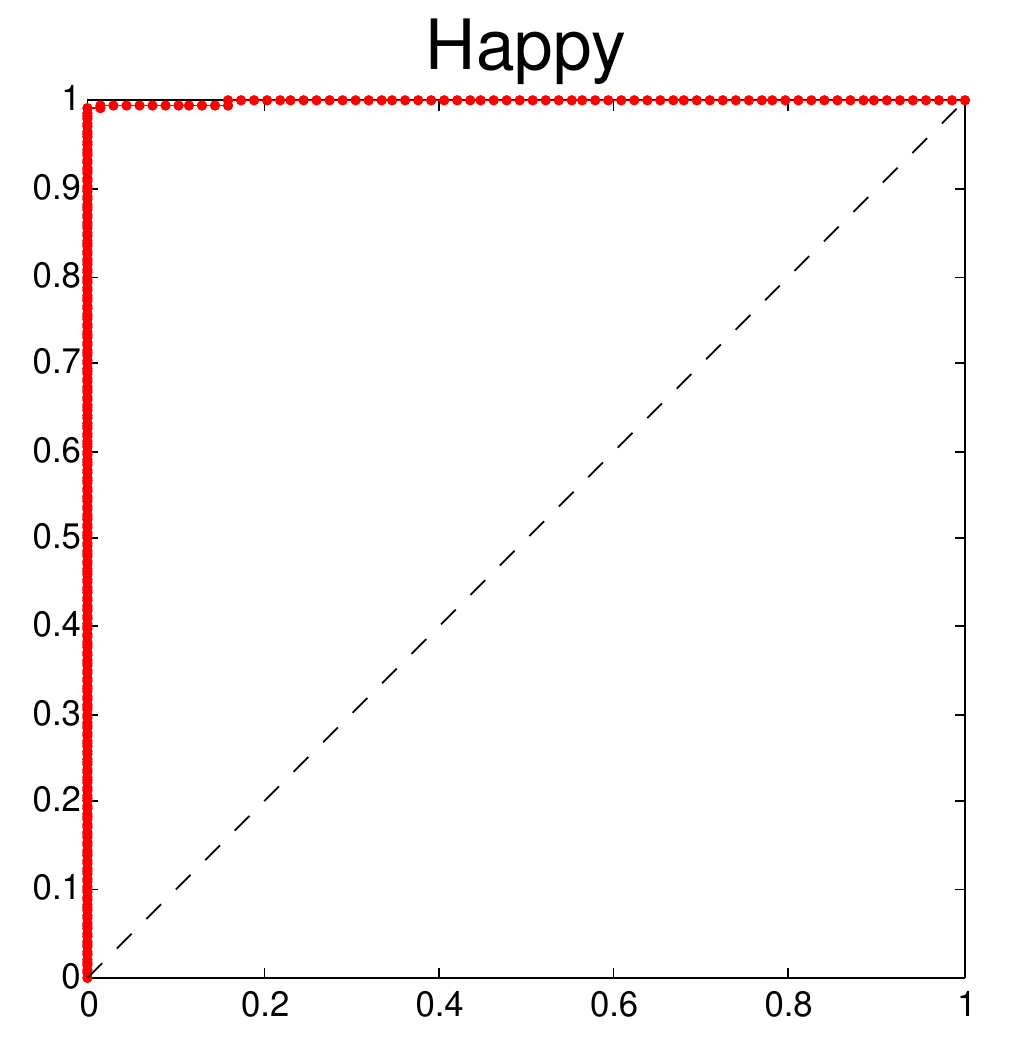} }  
  \subfloat[]
  {
  \includegraphics[width=0.32\textwidth]{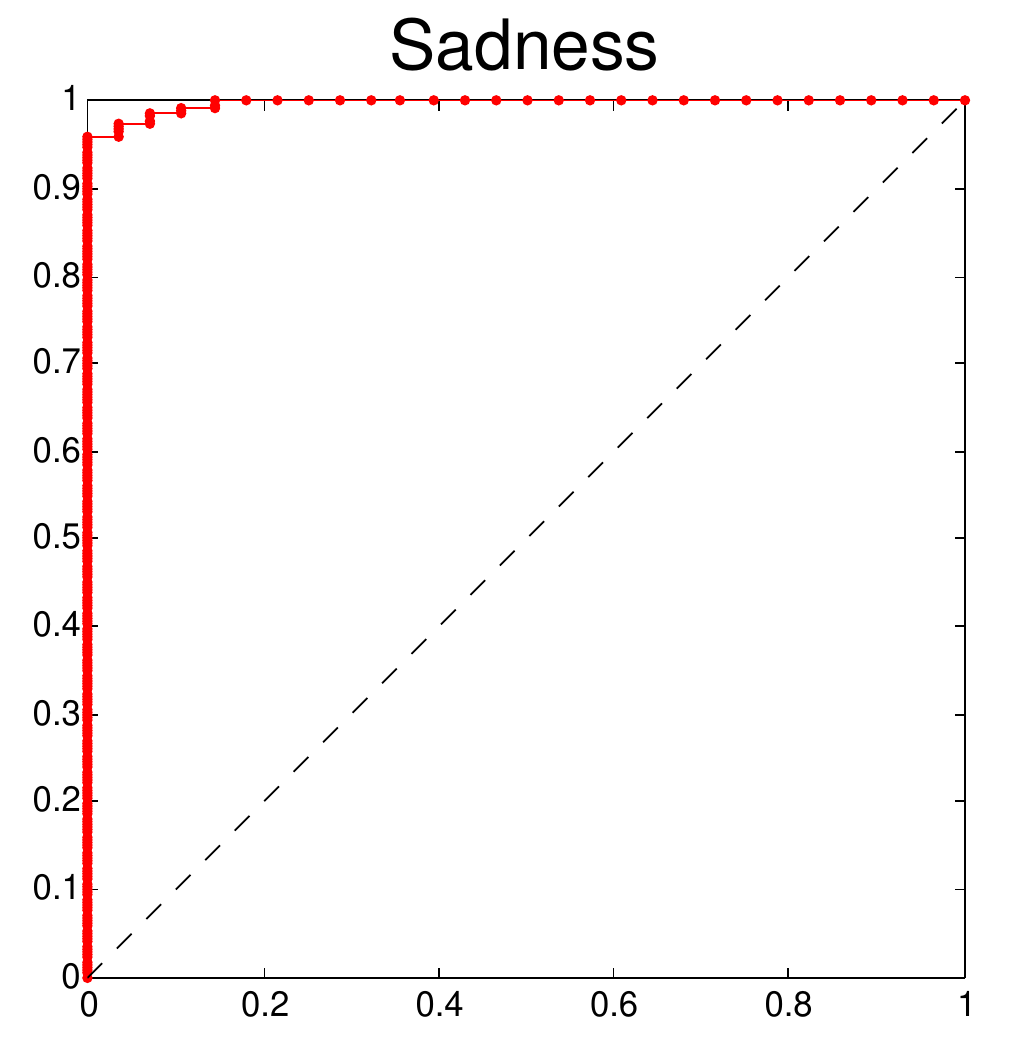} }  
  \subfloat[]
  {
  \includegraphics[width=0.32\textwidth]{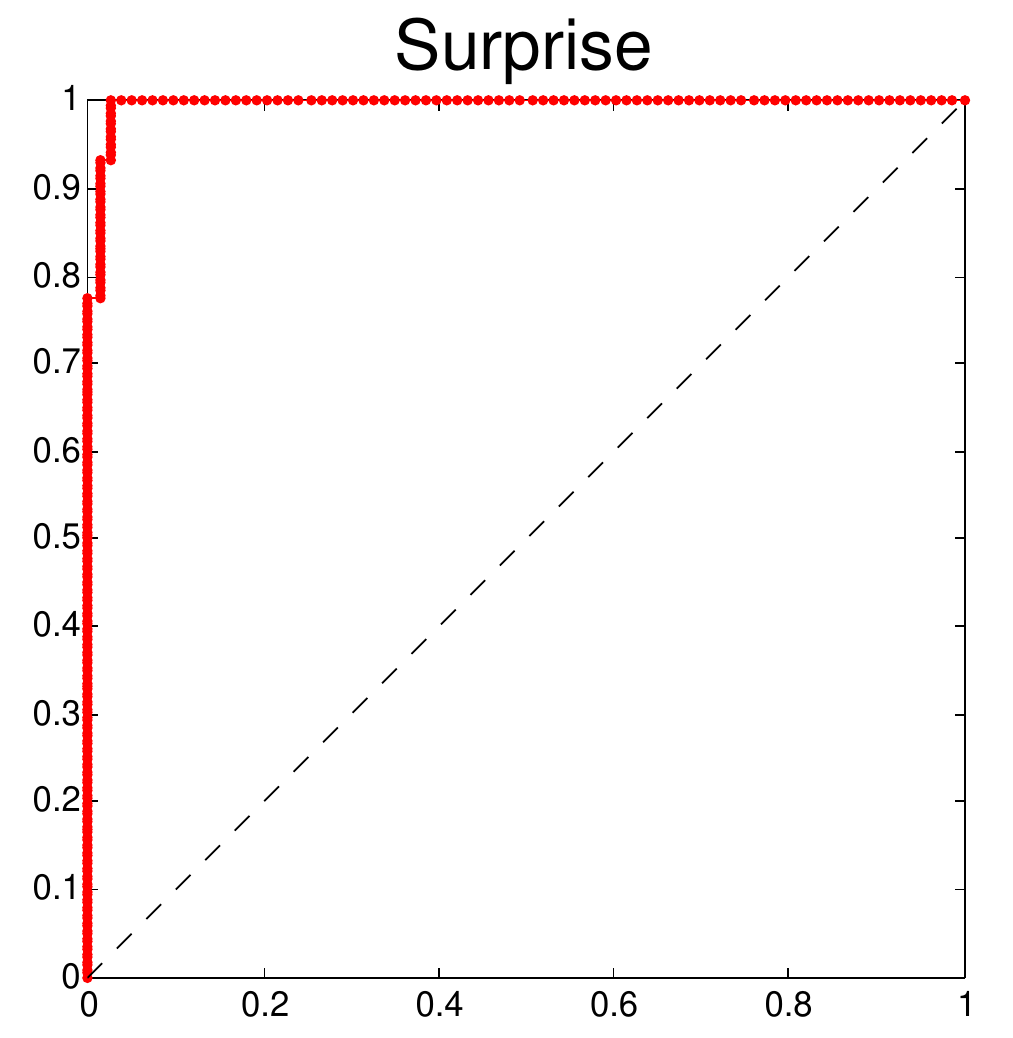} }    
  \caption{ROC curves of six emotion classifiers: (a) anger, (b) disgust, (c) fear, (d) happy, (e) sadness, (f) surprise. }
  \label{fig:roc}
\end{figure} 

We also compare our method with 4 best available results to date, including Time-series Kernels (Lorincz et al. \cite{lorincz2013emotional}), spatio-temporal independent component analysis (Long et al. \cite{long2012learning}), boosted dynamic features (Yang et al. \cite{yang2009boosting}), and non-negative matrix factorization techniques (Jeni et al. \cite{jeni2013continuous}). These methods are different from the baseline method in \cite{lucey2010extended}, because contempt emotion is removed from the data set, and binary classification is performed for each expression, so for the classification of one expression, the other five expressions are negative samples. Following the same experiment settings, the result of classification is shown in Figure \ref{fig:roc}, and detailed comparison is in Table \ref{Table: ComparisonROC}. The classification performance of our algorithm is close to, or even better than state-of-the-art method, with a better average accuracy. This demonstrate the effectiveness of the proposed LRBM model as both a binary classifier and a multi-class classifier.

\begin{table}
\begin{center}
\caption{AUC values of LRBM and the competing methods:  Time-series Kernels (Lorincz et al. \cite{lorincz2013emotional}), spatio-temporal independent component analysis (Long et al. \cite{long2012learning}), boosted dynamic features (Yang et al. \cite{yang2009boosting}), and non-negative matrix factorization techniques (Jeni et al. \cite{jeni2013continuous})}
\label{Table: ComparisonROC}
\small
\begin{tabular}{|c|c|c|c|c|c|c|c|}
\hline
Method & An & Di & Fe & Ha & Sa & Su & Avg. \\
\hline\hline
Yang et al. \cite{yang2009boosting} & 0.973 & 0.941 & 0.916 & 0.991 & 0.978 & 0.998 & 0.966\\
Long et al. \cite{long2012learning} & 0.933 & 0.988 & 0.964 & 0.993 & 0.991 & \textbf{0.999} & 0.978\\
Jeni et al. \cite{jeni2013continuous} & 0.989 & \textbf{0.998} & 0.977 & 0.998 & 0.994 & 0.994 & 0.992\\
Lorincz et al. \cite{lorincz2013emotional} & 0.991 & 0.994  & 0.987  & \textbf{0.999}  & 0.995  & 0.996  & 0.994 \\\hline
LRBM & \textbf{0.992}  & 0.995 & \textbf{0.995}  & \textbf{0.999}  & \textbf{0.997}  & 0.995  & \textbf{0.996} \\\hline
\end{tabular}
\end{center}
\end{table}

\subsubsection{Comparison with Feature Learning Method}

To comprehensively evaluate our method, we compare with a feature learning method, since RBMs are typically used for feature learning. A single RBM is trained on all classes of sequences using contrastive divergence. The size of hidden layer is the same as in the generative model. In the test process, given each observation sequence $\textbf{V}$, we compute the posterior probability $P(h_j|\textbf{V})$ for each hidden variable $h_j$ according to Equation. \ref{eqn: prob hidden}, as the feature representation. This is reasonable because $P(h_j=1|\textbf{V})$ is the probability of a pattern being activated. A linear multi-class SVM is trained on such features for classification.

The confusion matrix of this method is given in Table \ref{Table:resultfea}. For most expressions, the feature-based method is not as good as the generative model. The average recognition accuracy is 83.8\%, which is approximately 5\% below the performance of the generative model. This is reasonable, because the features learned from the model cannot capture the spatial relationship in each frame, which is specifically modeled in our model using the matrix $U$. If we add another latent layer, the classification performance only increases marginally by about 1\%.

\begin{table}
\begin{center}
\caption{Confusion Matrix of the classification performance using the feature learning method.}
\label{Table:resultfea}
\begin{tabular}{|c|c|c|c|c|c|c|c|}
\hline
 & An & Di & Fe & Ha & Sa & Su & Co \\
\hline
An & 88.9 & 11.1  & 0.0  & 0.0  & 0.0  & 0.0  & 0.0 \\
Di & 11.9  & 86.4 & 0.0  & 1.7  & 0.0  & 0.0  & 0.0 \\
Fe & 0.0  & 4.0  & 36.0 & 12.0  & 0.0  & 48.0  & 0.0 \\
Ha & 0.0  & 0.0  & 1.4  & 97.1 & 0.0  & 0.0  & 1.4 \\
Sa & 17.9  & 0.0  & 3.6  & 0.0  & 57.1 & 10.7  & 10.7 \\
Su & 0.0 & 2.4  & 0.0  & 0.0  & 0.0  & 97.6 & 0.0 \\
Co & 5.6  & 5.6  & 5.6  & 27.8  & 5.6  & 5.6  & 44.4 \\
\hline
\end{tabular}
\end{center}
\end{table}

\subsection{Human Action Recognition}

G3D data set is an action data set containing a range of gaming actions captured by Microsoft Kinect. The data set contains 10 subjects performing 20 gaming actions: \textit{punch right, punch left, kick right, kick  left, defend, golf swing, tennis swing forehand, tennis swing backhand, tennis serve, throw bowling ball, aim and fire gun, walk, run, jump, climb, crouch, steer a car, wave, flap and clap}. Synchronized video, depth and skeleton data are available in this data set. We only pick the skeleton data. To reduce the data dimension but keep the useful information, we use the 3D location information of four dominant joints (i.e. two hands and two feet). However the proposed approach can be applied to modeling more joints if we have enough training data.


Before abstracting the features, we perform a normalization procedure to minimize the effect brought by difference in subjects' body shapes. Specifically, we obtain the average bone lengths from all subjects in the training data, and then change the skeleton tracking results in every frame for each subject with the average shape. Therefore, every subject has the same body shape, and the cross subject distinction is alleviated.

\begin{figure}
\begin{center}
   \includegraphics[width=0.9\linewidth]{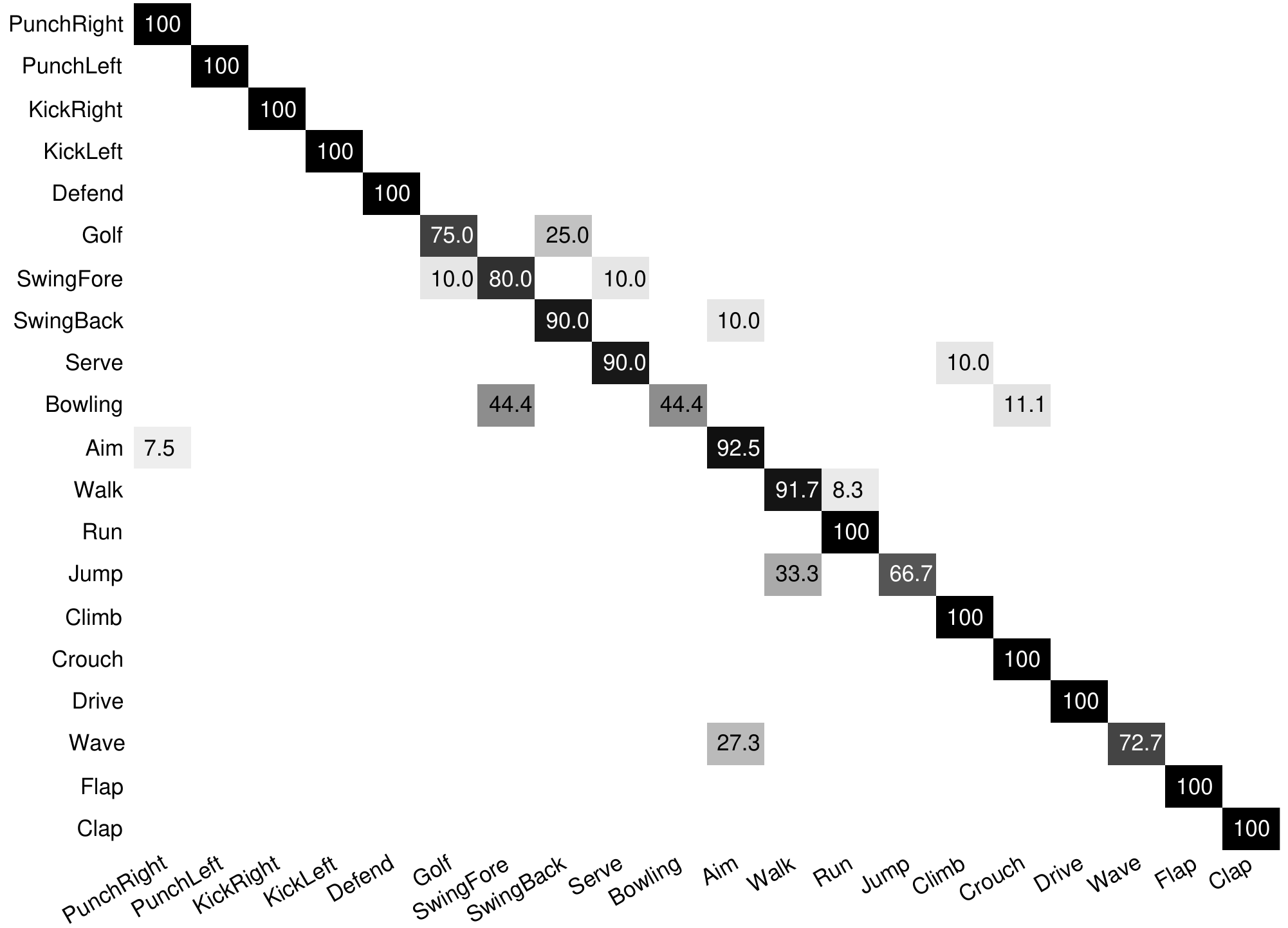}
\end{center}
   \caption{The confusion matrix of the proposed method on G3D data set}
\label{fig:G3DCM}
\end{figure}

As the size of the visible layer of an RBM is fixed, linear interpolation is performed to convert all sequences into the same length (20 frames for each sequence).  The 3D positions of the body joints along all three dimensions (x, y and z) form the 240-dimension input for the LRBM model. In the training phase, we set the size of hidden layer as 80.

We use the action segmentation that is provided by the data and the same experiment configuration as in \cite{bloom2012g3d}. The data set is split by subjects where the first 4 subjects were used for training, 1 for validation and the remaining 5 subjects for testing. 

\subsubsection{Classification Performance}

The confusion matrix is given in Figure \ref{fig:G3DCM}. The overall accuracy of LRBM is 90.5\%. Our model encounters some failures for the actions of \textit{ThrowBowlingBall} in \textit{Bowling} category and \textit{Jump}. In these actions, occasionally the body parts are occluded from the Kinect sensor, which will lead to estimated positions of joints. Then the positions of the two feet are significantly corrupted. As our method depends fully on the trajectories of joints, corrupted tracking results have a huge influence on the performance of LRBM. If we remove the pairwise potential in the visible layer and make the model a standard RBM, the accuracy drops to 84\%, which proves the better performance of LRBM than RBM, as expected, due to modeling the spatial interactions. Comparison with the baseline model \cite{bloom2012g3d} is given in Table \ref{Table: F1}.  
Other than \textit{Bowling}, performance on \textit{golf} action is quite close to the baseline method, and on all the other actions, our approach outperforms the baseline method. The overall accuracy is increased by 15.5\% in terms of F1 score. 

One reason of the improvement is that the method in \cite{bloom2012g3d} is based on 3 frames, hence can only capture local dynamics, while the proposed method is sequence based that can handle global dynamics.
\begin{table}
\begin{center}
\caption{F1 score of LRBM and baseline model}
\label{Table: F1}
\begin{tabular}{|c|c|c|}
\hline
Action & Bloom et al. \cite{bloom2012g3d} & LRBM\\
\hline\hline
Fighting & 70.46  & \textbf{97.09}\\\hline
Golf & \textbf{83.37}  & 81.82\\\hline
Tennis & 56.44  & \textbf{84.38}\\\hline
Bowling & \textbf{80.78}  & 61.54\\\hline
FPS & 53.57  & \textbf{95.52}\\\hline
Driving & 84.24  & \textbf{100.00}\\\hline
Misc & 78.21  & \textbf{95.24}\\\hline
Avg. & 72.44  & \textbf{87.94}\\\hline
\end{tabular}
\end{center}
\end{table}

To further demonstrate the effectiveness of global dynamic model over local dynamic model, we implement a conditional RBM \cite{taylor2007modeling} and a hidden Markov model for action recognition. For the conditional RBM model, each frame depends on 3 previous frames. 20 models are trained for 20 actions. Classification is based on the likelihood of a target sequence on different models. Basically the local dynamic model (CRBM or HMM) models one frame at a time and the global model (RBM or LRBM) models all frames at the same time. 

Similar to the expression recognition case, the posterior probability of the hidden variables $P(h_j|\textbf{V})$ can be used as the features for classification. We implement the feature-based method using linear SVM as the classifier. Again, it does not perform well compare with other dynamic methods, mainly due to the reason that it cannot capture the local spatial patterns in each frame. Details are given in Table \ref{Table: ACC}.

\begin{table}
\begin{center}
\caption{Recognition accuracy of different dynamic models. The feature learning method is denoted as $n$-FL-SVM for the feature learning nature and SVM classifier, with $n$ hidden layers.}
\label{Table: ACC}
\begin{tabular}{|c|c|}
\hline
Method & Accuracy (\%)\\
\hline\hline
HMM & 70.3\\\hline
1-FL-SVM & 75.9\\\hline
2-FL-SVM & 76.7\\\hline
CRBM \cite{taylor2007modeling} & 83.2\\\hline
RBM & 84.0\\\hline
LRBM & 90.5\\\hline
\end{tabular}
\end{center}
\end{table}

\subsubsection{Handling Noisy and Missing Data}

One advantage of generative model is that they can handle noisy or missing data in the input. To demonstrate this point, we design two scenarios with randomly selected noisy or missing data: 

(a) Noisy data. For a randomly selected portion of data, we multiply the ground truth data by a Gaussian noise $1+\mathcal{N}(0,1)$, where $\mathcal{N}(0,1)$ is the normal distribution; 

(b) Missing data. For a randomly selected portion of data, we assume they are missing, and use the average of their neighbors as the approximation when computing the likelihood. 

\begin{figure}
\begin{center}
   \includegraphics[width=0.7\linewidth]{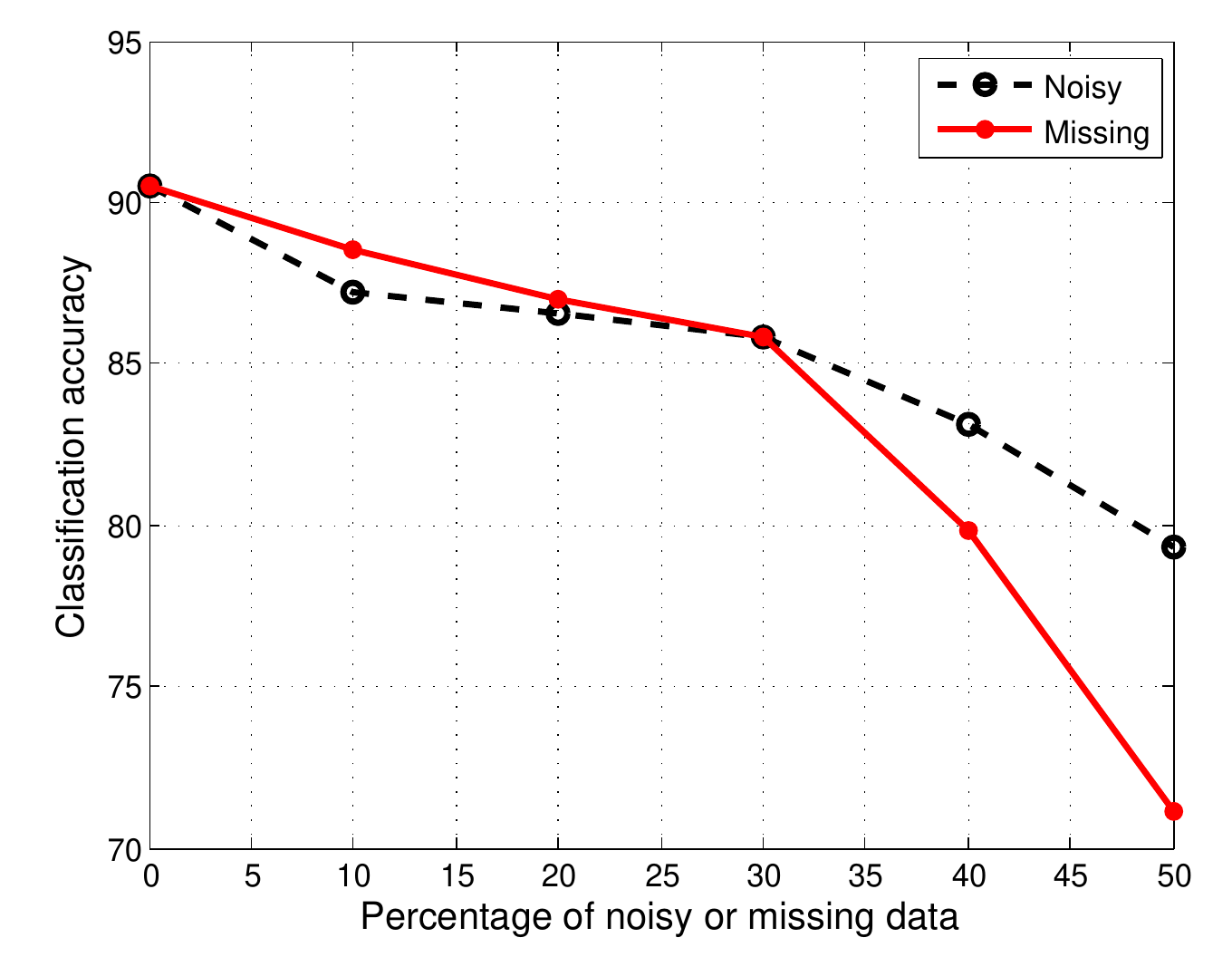}
\end{center}
   \caption{The classification performance with noisy or missing data.}
\label{fig:noisy}
\end{figure}

The data under these scenarios are then fed into our generative model for classification purpose. We vary the portion of noisy or missing data from 0 to 50\%, and observe the change of the classification accuracy. The details are given in Figure \ref{fig:noisy}. 

In scenario (a), with the portion of noisy data increase, the performance almost decays linearly. In scenario (b), if the percentage of missing data is small (less than 30\%), the average of its neighborhood is a decent estimation of the missing value, but with more data missing, the performance decays quite significantly. This is because several consecutive frames are missing at the same time, and the temporal information cannot be recovered effectively. One thing to notice is that with as much as 30\% noisy or missing data, our algorithm can achieve 86\% accuracy, only 4\% below the noiseless situation, which demonstrates the effectiveness of our algorithm to handle noisy or missing data.

\subsection{Complexity}

In the learning procedure of LRBM, we compute the data-dependent expectation and model-dependent expectation using matrix multiplication, so the computational complexity is linear with the size of the weight matrix ($dn_t\times n_h$) and iteration times ($n_{iter}$). Thus, the computation complexity is $O(dn_tn_hn_{iter})$. For facial expression recognition task, we set the epoch to 250, and the average running time for training one RBM is 1.77s, compared with 1.30s for standard RBM without local interaction. With 10 candidate models, it takes less than 5 min to train all the models. Experiments are performed on a desktop computer with an Intel i7 3.4GHz CPU and 8GB RAM. For testing, given a target sequence, we need to compute the unnormalized likelihood on each LRBM model using Equation \ref{eqn: detailed loglik}, with computational cost $O(n_td^2+n_hn_td)$. Given the unnormalized likelihood and the relative partition functions, the classification needs $n^2$ comparisons, therefore the overall classification complexity is $O(n(n_td^2+n_hn_td)+n^2)$, which is linear in the dimension of the input sequence. If we linearly increase the length of the sequence by increasing the frame rate, the complexity is also increased linearly.

For multi-class classification with $n$ classes, the confidence value matrix $S_\textbf{V}$ is built with complexity $O(n^2)$. In typical computer vision applications, the number of classes cannot exceed the dimension of the data, which means $n^2$ does not contribute much in the complexity, compared to the first term. To reduce the quadratic term $O(n^2)$ to a linear term $O(n)$, Annealed Importance Sampling \cite{salakhutdinov2008quantitative} can be used to estimate the partition function, thus the likelihood can be directly compared. However, the sampling method requires many samples to get a good estimation of the partition function, and the overall classification complexity is not reduced significantly.

Theoretically, the proposed method can be scaled up to data with thousands of dimensions, but much more training data will be needed since the number of parameters increases linearly with the size of input.

\section{Conclusion}
\label{conclusion}
In this paper, we study classification problem with multi-dimensional sequence data. To capture both the global dynamics and the local spatial interactions, we extend the conventional restricted Boltzmann machine by adding pairwise potentials in the energy function. An efficient mean field learning method is proposed to replace the reconstruction procedure in typical Contrastive Divergence learning, to estimate the additional parameters. Also, a novel label ranking approach of estimating the partition function of different LRBMs is presented to perform multi-class classification task. Experiments on two areas that involve multi-dimensional sequence data are performed to evaluate our approach. Results on two benchmark data sets prove the effectiveness of our algorithm.

Using the generative model for data representation, we do not need to deal with a deep and complicated structure, as in feature learning, and do not need to carefully tune the parameters. This is a major advantage of our method compared with feature learning methods. However, one shortcoming is that is that the data sequences have to be aligned for our global model to function properly, and the classification complexity is quadratic in the number of classes. We will address these issues in future work.

\section*{References}

\bibliography{mybibfile}

\end{document}